\documentclass[letterpaper, 10 pt, conference]{ieeeconf}
\IEEEoverridecommandlockouts
\usepackage{cite}
\usepackage{amsmath,amssymb,amsfonts}
\usepackage{algorithmic}
\usepackage{graphicx}
\usepackage{textcomp}
\usepackage{xcolor}
\usepackage{url}

\newif\ifarxiv
\arxivtrue
\usepackage{hyperref} 

\begin{document}
% ---- placeholder-figure helpers -----------------------------
% \phfig{height}  single-column empty figure box
% \phfigw{height} double-column empty figure box (use in figure*)
\newcommand{\phfig}[1]{%
  \fcolorbox{gray!60}{gray!8}{%
    \parbox[c][#1][c]{0.98\linewidth}{\centering
      \color{gray!70}\itshape [\,figure placeholder\,]}}}
\newcommand{\phfigw}[1]{%
  \fcolorbox{gray!60}{gray!8}{%
    \parbox[c][#1][c]{0.985\linewidth}{\centering
      \color{gray!70}\itshape [\,figure placeholder\,]}}}
% -------------------------------------------------------------

%\title{Zero-Shot sim2real Active Camera Control for Particle Image Velocimetry}
\title{Autonomous Particle Image Velocimetry Using Zero-Shot Sim-to-Real Reinforcement Learning}
\ifarxiv
\author{Chang Rajani$^{1}$, Luke Sebastien Marsden$^{1,2}$, Stephan Kunz$^{1}$, Claudio Mucignat$^{1}$, Ivan Lunati$^{1,2}$%
\thanks{$^{1}$Computational Engineering Lab, Empa, Swiss Federal Laboratories
for Materials Science and Technology, Switzerland.}%
\thanks{$^{2}$ETH Zurich, Switzerland.}%
\thanks{This work has been submitted to the IEEE for possible publication.
Copyright may be transferred without notice, after which this version may
no longer be accessible.}%
\thanks{Supplementary Video: \url{https://youtu.be/R1fhBA8eIFc}}%
}
\else
\author{Anonymous Authors}
\fi

\maketitle

\begin{abstract}
Particle image velocimetry (PIV) reconstructs a flow field from the motion of tracer particles but the reconstruction quality is bounded by how well the flow structure of interest is resolved in the images. The acquisition position is typically fixed by the operator before the measurement. We cast camera placement as a sequential decision-making problem and train a reinforcement learning agent to control a 6-DoF manipulator equipped with a camera, enabling it to locate, frame and resolve a hydrofoil wake in a water tunnel. The agent commands the camera pose and observes the scene only through the velocity field and uncertainty estimated from time-separated image pairs. It is trained entirely in a photorealistic simulation of our water tunnel, with wakes generated by Computational Fluid Dynamics (CFD) simulations of hydrofoils with varying profiles, positions and angles of attack. After zero-shot transfer to the real water tunnel, the policy reliably locates and frames the wakes of randomly placed hydrofoils. It resolves the wake at a higher resolution than a fixed camera view and with lower error in the measured flow statistics, in roughly half the time of a full scan and with far fewer image pairs.

\end{abstract}

\section{Introduction}

Particle image velocimetry (PIV) is an optical technique for measuring fluid velocity fields by tracking the displacement of tracer particles between successive images of a flow illuminated by a light sheet~\cite{raffel2018piv}. The quality of the resulting velocity estimates depends critically on the imaging conditions. Parameters such as magnification, camera distance, separation time, and depth of field determine the spatial resolution of the measurement and the particle displacement between exposures. Established PIV guidelines provide recommendations for selecting these parameters~\cite{keane1990optimization,raffel2018piv}, but tuning them to a particular flow typically requires several iterations and the judgement of an expert operator. 

In conventional practice, those parameters are selected based on prior knowledge or expectation of the velocity magnitude and the location of relevant flow features. This makes the measurement vulnerable to deviations from these expectations. When features of interest are compact, move across the field of view, or occur in unexpected locations, the quality of the recordings may be insufficient to allow accurate velocity reconstruction. No subsequent improvement in the PIV algorithm can recover information that was not adequately captured, resulting in suboptimal velocity estimates and inefficient use of measurement resources.

Robot-mounted PIV systems address this limitation by traversing a probe along a pre-planned path at close range and stitching the resulting image tiles together afterwards~\cite{jux2018robotic,michaux2018robopiv}. This approach provides detailed coverage of the entire measurement plane, but requires lengthy scans and devotes measurement time to regions that may not contain features of interest. In both conventional and robot-mounted PIV systems, camera placement is determined before measurement and is not adapted based on the acquired data. As a result, the camera may cover too little or too much of the region of interest.

In this work, we propose treating \emph{where the camera looks} as a controlled, closed-loop variable, in the spirit of \emph{active sensing}~\cite{bajcsy1988active}. We mount the camera on a 6-DoF manipulator observing a water tunnel and train a learned policy to reposition it so that an \emph{a priori} unknown flow structure is brought into a favourable imaging configuration: within the field of view and at a distance that allows it to be fully captured and sufficiently resolved. 
An appropriate representation of flow regions with low energy has long been of particular interest in applications such as wind energy~\cite{Barthelmie2006} and air quality~\cite{Luo2018}. In this work, we focus on hydrofoil wakes, which are known to exhibit large variations in shape and size with changes in angle of attack and flow speed~\cite{Anderson2017}.

\begin{figure*}[t]
  \centering
  \includegraphics[width=\linewidth]{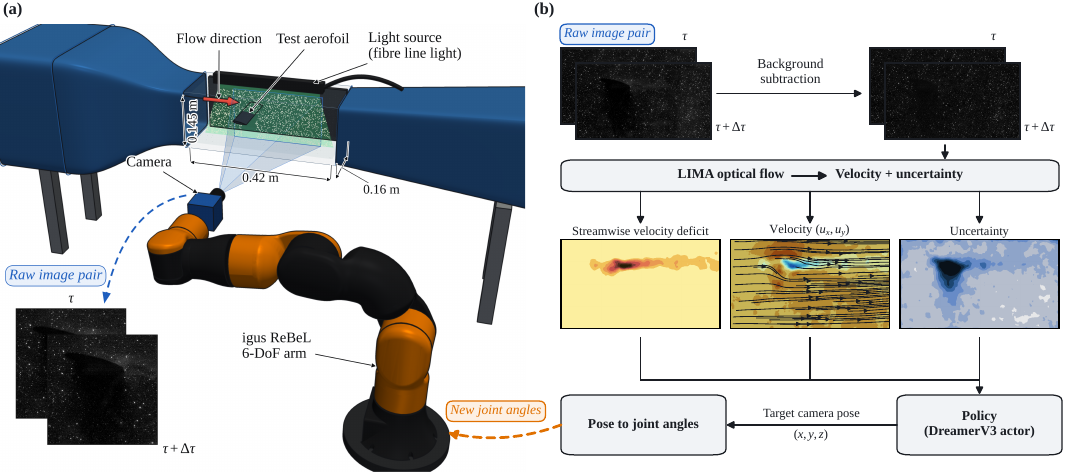}
\caption{\textbf{System overview.} (a) The water tunnel (blue, with a plexiglass test section in the middle) is seeded with tracer particles, and an LED line light illuminates a thin planar sheet of the flow. A test hydrofoil is placed at an arbitrary position and angle of attack and sheds a wake. A camera on the wrist of an igus ReBeL 6-DoF arm images the sheet. (b) At each decision step the camera records four image pairs at times $\tau$ and $\tau+\Delta \tau$. After background subtraction, LIMA estimates the velocity, from which we compute the streamwise velocity deficit, and its uncertainty. The policy takes these three fields as input and outputs a new camera position, which is converted to joint angles and sent to the arm.}
  \label{fig:system}
\end{figure*}

Casting camera placement as a sequential decision-making problem poses three challenges. First, the true flow field is not available to the agent, which can estimate the velocity only from its own camera images. These estimates are unreliable when the camera is badly positioned. Our agent therefore uses the flow estimate and uncertainty provided by LIMA, a CNN-based PIV estimator that is fast enough to run within the control loop~\cite{manickathan2023lima,mucignat2026plima}, to guide camera positioning decisions. Second, reinforcement learning requires millions of interactions, which would take weeks to collect on physical hardware. We train a DreamerV3 agent~\cite{hafner2025dreamer} entirely in a simulation of our laboratory setup built in NVIDIA Isaac Sim with Isaac Lab~\cite{mittal2023orbit}. The simulation includes the test section, light sheet, tracer particles and camera optics of the real setup, together with flow fields from CFD simulations. We then deploy the policy on the physical manipulator without fine-tuning or training on real-world data. Third, camera viewpoint selection and robot joint motion are defined in different spaces. The agent commands a Cartesian camera position, and a calibration of the manipulator against a marker board in the test section maps each command to joint angles (Fig.~\ref{fig:system}).

Our contributions are threefold: (i)~a system that formulates active PIV camera placement as a reinforcement learning problem over Cartesian camera-position commands, in which the policy must locate, frame, and resolve an initially unknown wake; (ii)~a high-fidelity Isaac Sim model of our laboratory PIV setup, incorporating CFD flow fields for different NACA hydrofoils to simulate tracer-particle motion; and (iii)~zero-shot deployment on physical hardware, demonstrated through extensive experiments showing that the policy consistently converges to camera positions that outperform a conventional fixed view and achieve flow-estimation quality comparable to brute-force image stitching at a substantially lower measurement time.

% ============================================================
\section{Related Work}
\textbf{Particle Image Velocimetry.}
PIV is a quantitative and non-intrusive technique for reconstructing flow fields at high spatial resolution~\cite{raffel2018piv}. The flow is seeded with neutrally buoyant tracer particles, which are illuminated by a light source within a plane. The scattered light is recorded at two or more instants separated by a known interval (the \textit{recording} phase). In the \textit{evaluation} phase the particle displacement between frames is estimated, traditionally with statistical methods, but in recent years also with neural networks~\cite{manickathan2023lima,lagemann2021raftpiv}. This stage requires accurate camera calibration to relate the pixel displacements to physical displacements. Post-processing finally validates the vectors and replaces erroneous ones, using statistical criteria~\cite{raffel2018piv} or user-defined masks for regions affected by reflections or lacking tracer particles.

\textbf{Optimal PIV data acquisition and processing in practice.}
In conventional PIV, the imaging parameters are fixed before the experiment following established design rules~\cite{keane1990optimization,raffel2018piv}, and the measurement uncertainty is quantified afterwards from the recorded images~\cite{sciacchitano2019uncertainty} without feeding back into the viewpoint selection process. Robot-mounted systems follow the same pattern: robotic volumetric PIV traverses a probe around a full-scale object and stitches the recorded sub-volumes in post-processing~\cite{jux2018robotic}, while industrial RoboPIV
installations distribute illumination and cameras over several robots
to cover a large domain~\cite{michaux2018robopiv}. In both cases, the
trajectory is planned in advance and the acquired data does not
influence camera positioning. Methods that improve accuracy from the captured images, such as iterative image deformation~\cite{Scarano2002}, act on the \textit{evaluation} rather than the \textit{recording} phase and cannot improve data acquisition itself.

\textbf{Learned flow estimation and virtual camera agents.}
We use LIMA, a lightweight matching network that enables real-time inference during measurements~\cite{manickathan2023lima,mucignat2026plima} and provides an uncertainty estimate for the instantaneous displacement fields, as the core PIV algorithm in our work. Among existing approaches, the pykitPIV framework is the most related to our work. It provides configurable virtual PIV experiments and allows an agent to be trained to move a virtual camera across a synthetic image to locate sources and
sinks~\cite{zdybal2025pykitpiv}. In contrast to pykitPIV, our framework models the camera and lens and allows camera movements in three dimensions. The distance between the sensor and the measurement plane determines the magnification and the resolvable displacement. Each commanded viewpoint must be reached by a robotic arm subject to physical constraints such as joint limits. Our simulation also provides photorealistic rendering, including physically realistic glare, shadows, occlusions, and reflections. Finally, we transfer the trained policy directly to real hardware. 

\textbf{Active perception and viewpoint control.}
Selecting one's own observations is the goal of \emph{active
perception}~\cite{bajcsy1988active} and next-best-view
planning~\cite{zeng2020viewplanning}. In these approaches, candidate views are scored by expected information gain over an explicit scene model~\cite{isler2016infogain} or through a learned policy~\cite{chen2024gennbv}. These methods aim to cover a static scene, using a persistent geometric model to evaluate candidate views. Our objective also involves coverage, but focuses on a single flow structure rather than the scene as a whole. The structure is not directly visible in the images, but it is identified through the velocity estimated from image pairs. The agent must locate the structure and frame it entirely at the best possible magnification. Our setting is also related to eye-in-hand visual servoing~\cite{chaumette2006servo}, in which camera motion is controlled from image features in closed loop. However, our task has no predefined target image or image feature to track, and the structure of interest must first be identified. Reinforcement learning in experimental fluid mechanics has generally been used in flow control~\cite{vignon2023advances} or navigation problems~\cite{singh2023plumes} rather than to improve observations of the flow.

\textbf{Sim-to-real transfer.}
Domain randomisation is the standard approach to improving policy robustness to differences between rendered and real images~\cite{tobin2017domainrand}. An alternative is to provide the policy with an intermediate representation that is largely invariant to appearance. This approach has been used to transfer driving policies from simulation to the real world~\cite{pmlr-v87-mueller18a}. Optical flow is one such representation, and it has been used for visual servoing on real robots because of this robustness~\cite{argus2020flowcontrol}. We combine the two approaches. The policy receives the output of the flow estimator as its observation, reducing the appearance differences between simulation and reality, while retaining errors introduced by the estimator itself. Where the light sheet is shadowed or unlit, the estimator may report spurious flow, and the policy must learn to ignore these regions (Sec.~\ref{sec:spurious}).

\section{Problem Formulation}
\label{sec:problem}

We mount a camera on a 6-DoF robotic arm in front of a water tunnel seeded with tracer particles. A vertical LED light sheet illuminates a thin streamwise slice of the test section, which defines the measurement plane, and a hydrofoil generates a wake in a uniform free stream (see Fig.~\ref{fig:system}). The position and extent of the wake depend on the hydrofoil geometry, position and angle of attack (AoA), and on the free-stream speed, none of which is known to the agent. The agent must \emph{search} for the wake by acquiring a sequence of velocimetry measurements, and then maintain a viewpoint from which the flow estimator can resolve the wake accurately. The hydrofoil may also be exchanged for another one at a different position, so the agent must detect that the tracked wake has moved or disappeared and search again.

\begin{figure}[t]
  \centering
  \includegraphics[width=\columnwidth]{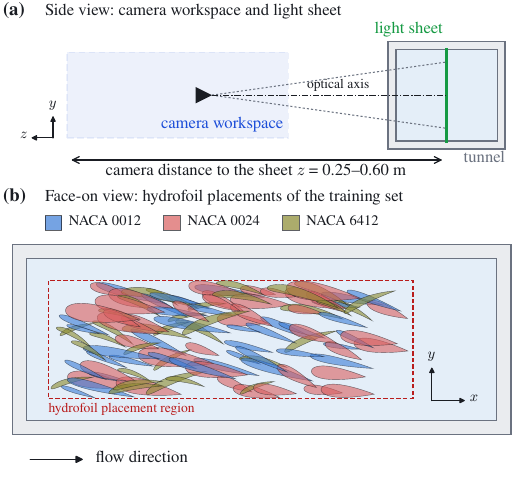}
  \caption{\textbf{Task geometry and training set.} (a) Side view. The agent commands the camera optical centre inside the camera workspace (dashed), with the optical axis held normal to the light sheet. (b) Face-on view of a subset of the hydrofoil placements of the training set, coloured by NACA section. The angle of attack, chord and section vary per placement, and training uses two free-stream speeds. Placements lie inside the dashed region, which keeps 20\,mm from the inlet, the floor and the lid, and 75\,mm from the downstream end so that the wake stays in the test section.}
  \label{fig:task}
\end{figure}

We formulate the problem as a partially observable Markov decision process (POMDP)~\cite{kaelbling1998pomdp} with a horizon of $T$ decision steps (we use $T=15$ or $T=45$ depending on the training phase, see Fig.~\ref{fig:sim_results}). The hydrofoil parameters, tracer positions, and arm configuration are unknown to the agent, which only observes the flow estimate that LIMA infers from the recorded frames~\cite{mucignat2026plima}. We operate in the standard online RL setting~\cite{sutton2018rl} and maximise the expected discounted return $\mathbb{E}_{\pi}\left[\sum_{t=1}^{\infty} \gamma^{t-1} r_t\right]$, with $\gamma = 0.997$. The horizon $T$ imposes a time limit but does not define a terminal state.

Throughout this work, we use $x$ for the streamwise coordinate, $y$ for the vertical coordinate, and $z$ for the distance of the camera from the sheet. The action $\mathbf{a}_t = (x_t, y_t, z_t)$ specifies the bounded absolute position of the camera optical centre within a Cartesian camera workspace $\mathcal{A}\subset\mathbb{R}^3$ of $0.30\,\mathrm{m} \times 0.135\,\mathrm{m} \times 0.35\,\mathrm{m}$, with its longest axis normal to the sheet. One action component sets the camera distance, $z_t \in [0.25, 0.60]$\,m, while the other two translate the imaged window across the sheet. In 2D PIV, the optical axis must remain normal to the imaging plane, so the agent cannot pitch, yaw, or roll the sensor. The arm is moved to the commanded position by inverse kinematics in simulation and by a calibrated lookup table on hardware (Sec.~\ref{sec:method}). If the arm cannot realise the command, the agent receives a reward of $r_t=-2$.

The reward is computed from the time-averaged velocity field $\mathbf{u}(\mathbf{x})\equiv(u_x,u_y)$ of the CFD case, where $\mathbf{x}\equiv(x,y)$ denotes a point on the measurement plane. Let $U$ denote the free-stream speed of the episode. The \emph{streamwise velocity deficit} is defined as
\begin{equation}
  \delta(\mathbf{x}) \;=\; \frac{U - u_x(\mathbf{x})}{U},
  \label{eq:deficit}
\end{equation}
which is near zero in the free stream and one where the flow is stationary. The \emph{wake region} $\mathcal{S}$ is defined as the part of the sheet connected to the hydrofoil where the deficit exceeds a threshold that increases with the distance $s(\mathbf{x})$ from the hydrofoil surface,
\begin{equation}
  \delta(\mathbf{x}) \;>\; \delta_0 \left(1 + \frac{s(\mathbf{x})^{2}}{\lambda^{2}}\right),
  \label{eq:region}
\end{equation}
with $\delta_0 = 0.05$ and $\lambda = 100$\,mm. The low threshold near the hydrofoil surface ensures that thin wakes at small angles of attack are included in the region, while the increasing threshold excludes regions where the wake has decayed. Slow-flow regions that are not connected to the hydrofoil, such as the boundary layers along the test-section walls, are excluded from $\mathcal{S}$.

At step $t$, the camera faces the sheet at distance $z_t$. With an image resolution of $W\times H$ pixels and focal length $f$, it captures a window $\mathcal{W}_t$ of the sheet with area $A_t = W H\, z_t^{2} / f^2$. A point in $\mathcal{S}$ is considered in view if it projects onto the image and is not occluded by the hydrofoil. We define \emph{capture} as the fraction of the wake that is in view and \emph{fill} as the fraction of the frame occupied by the wake,
\begin{equation}
  c_t \;=\; \frac{|\mathcal{S} \cap \mathcal{W}_t|}{|\mathcal{S}|},
  \qquad
  \phi_t \;=\; \frac{|\mathcal{S} \cap \mathcal{W}_t|}{A_t}.
  \label{eq:capture}
\end{equation}
Fill decreases as $1/z_t^{2}$, so the closest pose that captures the entire wake achieves the highest fill. This is the \emph{required pose} $\mathbf{p}^{*}_{\mathcal{S}} = (x^{*}, y^{*}, z^{*})$ for a given case (as the orientation of the camera is fixed, poses are fully defined by Cartesian coordinates). The required pose is centred on the bounding box of $\mathcal{S}$ and is located at the smallest distance at which the window spans the entire box,
\begin{equation}
  z^{*} \;=\; \max\!\left(\frac{f\, \Delta x_{\mathcal{S}}}{W},\; \frac{f\, \Delta y_{\mathcal{S}}}{H},\; z_{\min}\right),
  \label{eq:required}
\end{equation}
where $\Delta x_{\mathcal{S}}$ and $\Delta y_{\mathcal{S}}$ are the dimensions of the bounding box, and $z_{\min} = 0.25$\,m is the near wall of the camera workspace. If the bounding-box centre lies outside the camera workspace, the required pose is placed at the nearest reachable position, and $z^{*}$ is increased until the wake fits within the image window. We denote the capture and fill of the required pose by $c^{*}_{\mathcal{S}}$ and $\phi^{*}_{\mathcal{S}}$, respectively. The capture $c^{*}_{\mathcal{S}}$ is slightly below one because the hydrofoil occludes part of the wake near its surface.

The per-step reward is defined as
\begin{equation}
  r_t \;=\; \underbrace{\mathrm{clip}\!\left(\frac{c_t - c_0}{c^{*}_{\mathcal{S}} - c_0},\,0,\,1\right)}_{\text{capture}}
        \,\underbrace{\min\!\left(\frac{\phi_t}{\phi^{*}_{\mathcal{S}}},\,1\right)}_{\text{fill}},
  \label{eq:reward}
\end{equation}
with $c_0 = 0.6$. Both factors attain their maximum value of one at the required pose, so the reward is one regardless of the wake size. The same pose serves as the reference for evaluating the policy in Sec.~\ref{sec:sim}. In our experiments without the normalisation by $\phi^{*}_{\mathcal{S}}$, the agent preferred small wakes because they gave a higher fill for the same quality of framing. Without clipping at one, the agent moved closer than the required pose and cropped the wake to increase the fill.

Between the far and the near wall of the camera workspace, the magnification on the sheet changes by a factor of 2.4, so a wake that fills the frame is sampled at more than twice the spatial resolution of one that occupies only a corner. An ideal agent must therefore \emph{search} for the wake (make $c_t$ nonzero), \emph{frame} it (increase $c_t$ towards $c^{*}_{\mathcal{S}}$), and \emph{resolve} it (move closer to $z^{*}$). Since $z^{*}$ varies from case to case, the agent must position the camera closer to the test section for small wakes and further away for large ones, rather than memorising a single suitable position.

As the wake fluctuates in time, PIV systems commonly average over several shedding cycles to obtain a mean flow field. Therefore, to save simulation time, we advect the tracer particles using the time-averaged velocity field from CFD. At step $t$, once the arm has settled, the camera records four frame pairs. The two frames in each pair are taken with an inter-frame time $\Delta \tau = 10$~ms. LIMA estimates a displacement field and its uncertainty from each pair; both are averaged over the four pairs. The calibrated camera model converts the averaged displacement and uncertainty into a velocity estimate $\hat{\mathbf{u}}$ and uncertainty $\hat{\sigma}$, respectively. Using the known free-stream speed, we obtain the estimated deficit $\hat{\delta}$ from Eq.~\eqref{eq:deficit}. The observation is $o_t = (\hat{\mathbf{u}}, \hat{\delta}, \hat{\sigma}/U)$\footnote{For clarity, we drop the hat notation in the experiments since all quantities there are estimated. 
}. The uncertainty is high in regions where tracers are not illuminated, allowing the agent to distinguish the wake from poorly lit regions (Sec.~\ref{sec:spurious}). We use four pairs as a compromise between rendering time and the quality of the flow estimate required to locate and frame the wake. The reward is a privileged signal available only in simulation; the deployed policy relies solely on $o_t$, enabling zero-shot transfer (see Sec.~\ref{sec:sim2real}). 
% ============================================================
\section{System and Method}
\label{sec:method}

\subsection{Camera calibration}
\label{sec:calib}
The policy commands a view, defined by the point of the light sheet at the image centre and the distance $z$, and the arm must position the camera accordingly. On hardware, a rigid camera-mount calibration combined with the arm kinematics produced pose-dependent view errors. We therefore calibrate the view directly. A large ArUco board is held in the water in the plane of the light sheet, and the arm visits a set of poses distributed over the camera workspace, all in the same wrist configuration. In each image the detected markers provide a homography between the image and the board, from which we extract the position of the image centre on the sheet, the pixel size on the sheet, and the image roll. These quantities are stored together with the joint angles reported by the arm. Since the board is imaged through the glass and the water, refraction is included in the measurement and requires no explicit model. To reach a commanded view, we interpolate the table to obtain the joint angles and execute a single joint move. For views not contained in the table, the image centre lands within about 2\,mm of the commanded point. Commands outside the calibrated region are refused, and the arm does not move. During training, we displace the camera by a random error of a few millimetres in every episode, so that the policy does not rely on exact command execution.

\subsection{World-model agent}
The policy is trained using DreamerV3~\cite{hafner2025dreamer} without modifications. DreamerV3 fits a recurrent latent world model to the observations, while the actor and critic are optimised using imagined rollouts. Model-based RL is well suited to this setting because each environment step requires computationally expensive path-traced rendering in Isaac Sim, making sample efficiency particularly important.

\subsection{Simulation and rendering}
\label{sec:sim_rendering}

\begin{figure}[h]
  \centering
  \includegraphics[width=\columnwidth]{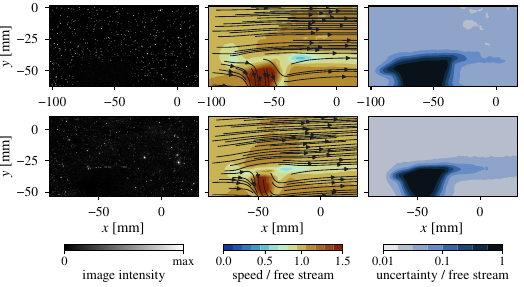}
  \caption{\textbf{Simulated (top row) and real observations (bottom row).} A simulated and a real hydrofoil of similar profile at a similar angle of attack, shown for illustration. The two rows show different fields of view. \textbf{Left:} one camera image after preprocessing. \textbf{Middle:} flow speed relative to the free stream, averaged over 100 image pairs. \textbf{Right:} LIMA uncertainty relative to the free-stream speed. In both rows the wake is a slow band behind the trailing edge. Below the hydrofoil, the estimator returns a spurious flow in the shadow.}
  \label{fig:sim_obs_real}
\end{figure}

We build the training environment in NVIDIA Isaac Sim to reproduce the laboratory setup. The simulated arm, imported from its URDF, has exact kinematics and therefore does not need a calibration table. We place it at the inverse-kinematics solution of each commanded view, and the residual error of the real calibration is modelled by the random camera offset described in Sec.~\ref{sec:calib}. The motion between viewpoints is not simulated because it carries no information for the task. The tunnel has the real inner dimensions and is filled to the brim (see Fig.~\ref{fig:system}). The camera is modelled as a pinhole with the calibrated intrinsics of the real sensor and lens, mounted at a calibrated offset from the flange. Its output is convolved with a diffraction-limited point spread function (PSF). The PSF width is scaled and tracers are rendered larger than their real size, so that the rendered particle images roughly resemble those captured on hardware. We omit depth of field because the motorised lens is refocused on the sheet at every camera distance, keeping the sheet in focus.

The flow in each episode is a precomputed two-dimensional lattice-Boltzmann solution for the hydrofoil in a domain that approximates the test section of the tunnel. It is computed with XLB~\cite{ataei2024xlb} on a D2Q9 lattice with the entropic KBC collision model~\cite{bosch2015kbc}, at a grid spacing of 0.4\,mm, and averaged over 24\,s of shedding. We train on 719 cases from 372 hydrofoil configurations at two free-stream speeds. The configurations span three NACA sections (0012, 0024 and 6412), angles of attack of up to $30^\circ$ in either direction, chord lengths of 30 to 70\,mm, and different positions in the test section (see Fig.~\ref{fig:task}). Each episode draws one case and randomly mirrors it left to right or top to bottom. These transformations are exact because the test section is symmetric and the two-dimensional flow is unaffected by gravity.

Each environment is seeded with tracers in the light-sheet plane, which are advected through the time-averaged CFD field over the 10\,ms inter-frame interval of the real camera. We render the scene before and after the advection. Figure~\ref{fig:sim_obs_real} compares the images supplied to LIMA and the resulting velocity fields and uncertainties for the real water tunnel and the simulation of a similar hydrofoil condition. The tracers are the only bright objects and are illuminated from above by an LED line light that forms the sheet. Illumination is computed by the Isaac Sim path tracer, which naturally reproduces effects such as shadows cast by the hydrofoil through the particle field and occlusion of the sheet from oblique viewpoints. The hydrofoil can be replaced by the CAD model of any other object without making any change to the rendering code. 

LIMA tends to report higher uncertainty in a real turbulent wake than in the free stream, due to increased particle loss from the flow being more three-dimensional. To reproduce this effect in simulation with the smooth mean CFD field, we add a small, random, out-of-plane velocity vector to each tracer, with a magnitude that increases with the local velocity deficit. We randomise the seeding density, tracer-size distribution, and sensor noise in every episode to cover conditions that are difficult to measure directly.

\subsection{Spurious deficits}
\label{sec:spurious}
The estimated deficit $\hat{\delta}$ is high wherever the flow is slow, but also where the flow estimator has no lit tracers to track. This occurs in the hydrofoil shadow, in the upper corners of the test section beyond the reach of the LED line light, and near the walls. In the shadow, for example, the estimator returns a streamwise velocity well below the free-stream speed, together with a spurious downward velocity, while its uncertainty is much higher than that of the free stream (Fig.~\ref{fig:sim_obs_real}). Such regions can be larger than the wake itself. 

We initially tested a scripted approach in simulation: two wide views survey the test section, Eq.~\eqref{eq:region} is applied to the estimated deficit, and the camera moves to the required pose defined by Eq.~\eqref{eq:required}. Even with manually generated masks based on the tunnel and light geometry, this approach often framed the shadow instead of the wake. We therefore render the light, shadow, and tracers physically, reproducing the same spurious deficits during training, and provide the policy with the LIMA uncertainty estimate~\cite{mucignat2026plima}. Policies trained without uncertainty often framed the hydrofoil shadow or the unlit corners, particularly on hardware. With uncertainty, the policy learns to distinguish the wake from poorly lit regions without requiring manually generated masks.

\section{Simulation Results}
\label{sec:sim}

\begin{figure}[t]
  \centering
  \includegraphics[width=\columnwidth]{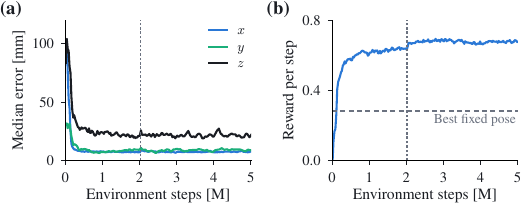}
  \caption{\textbf{Training in simulation.} The policy is trained for 2.0\,M environment steps on 15-step episodes with a fixed hydrofoil, then for 3.0\,M steps on 45-step episodes, half of which contain a hydrofoil change. The dotted line marks the switch. \textbf{(a)}~Median error between the settled and the required camera pose in $x$, $y$ and $z$. \textbf{(b)}~Reward per decision step. The dashed line is the best single fixed camera pose.}
  \label{fig:sim_results}
\end{figure}

\begin{figure}[t]
  \centering
  \includegraphics[width=\columnwidth]{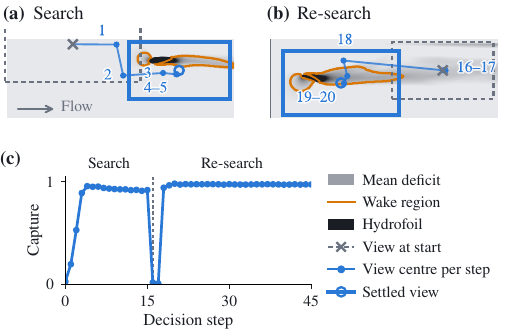}
    \caption{\textbf{Search and re-search in one evaluation episode.} The policy finds a wake within three steps, holds it, and after the hydrofoil is exchanged at step 16 it finds the new, larger wake within two steps and moves back to fit it. \textbf{(a, b)}~Face-on view of the test section before and after the exchange, with the time-averaged deficit (grey), the wake region $\mathcal{S}$ (orange), the path of the view centre numbered by decision step, and the settled view (blue rectangle). \textbf{(c)}~Capture $c_t$ (Eq.~\eqref{eq:capture}) per decision step. The policy settles at 373 and 425\,mm, against required distances of 350 and 415\,mm.}
  \label{fig:sim_episode}
\end{figure}

\begin{figure}[t]
  \centering
  \includegraphics[width=\columnwidth]{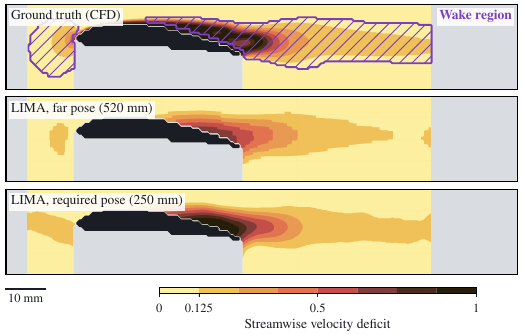}
  \caption{\textbf{Framing sharpens the measured wake.} From the required pose LIMA resolves the downstream part of the wake and the thin layer along the upper surface of the hydrofoil, neither of which are captured well at the far pose. Top: true deficit of one simulated wake, with the wake region $\mathcal{S}$ (purple). Middle and bottom: the deficit LIMA measures from 520\,mm and from the required pose at 250\,mm, at the same view centre, each averaged over 100 frame pairs. The hydrofoil is black, and its shadow (grey) is left out of the comparison.
  }
  \label{fig:lima_accuracy}
\end{figure}

We train the policy in two phases (Fig.~\ref{fig:sim_results}). In the first, episodes are 15 steps long and the hydrofoil remains fixed, allowing the policy to learn to find and hold a wake. In the second, the episodes are 45 steps long, and the hydrofoil is exchanged for another one midway through half of the episodes. The policy must therefore detect when the wake disappears and locate the new one. At the end of training, the median error between the settled and the required view is below 10\,mm in $x$ and $y$ and about 20\,mm in $z$. The distance is the hardest to estimate because the extent of a thin wake is difficult to infer from tracer particles. The per-step reward is more than twice that of the best fixed camera position and remains at the same level when hydrofoil changes are introduced.

Figure~\ref{fig:sim_episode} shows a representative evaluation episode. The wake is initially out of sight. The policy finds it within three steps and maintains the view. When the hydrofoil is exchanged at step 16, the wake disappears from view. The policy searches again and settles on the new, larger wake, which requires a greater camera distance. Across evaluation episodes, the policy captures the wake within a median of four steps from a random initial view and three steps after a hydrofoil change. Following a change, it frames the new wake as accurately as it does from a fresh start.

Figure~\ref{fig:lima_accuracy} illustrates the benefit of reaching the required pose. We measure the same simulated wake from 520\,mm and from the required pose, with the camera centred on the same point. From the farther pose, LIMA resolves the wake directly behind the trailing edge, but not its weaker downstream part or the thin boundary layer and separation region along the hydrofoil upper surface. These features are much better resolved from the required pose. The improvement results from spatial sampling: at the required pose, each millimetre of the wake spans about twice as many pixels, allowing LIMA to better resolve gradients at the wake boundary. Across the simulated wakes tested, LIMA identifies, on average, 54\% of the true wake region from the far pose, compared with 72\% from the required pose. The required pose performs better for nearly every wake, and the peak deficit is also closer to the true value.

\section{Sim-to-Real Transfer}
\label{sec:sim2real}

\begin{figure*}[t]
  \centering
  \includegraphics[width=\linewidth]{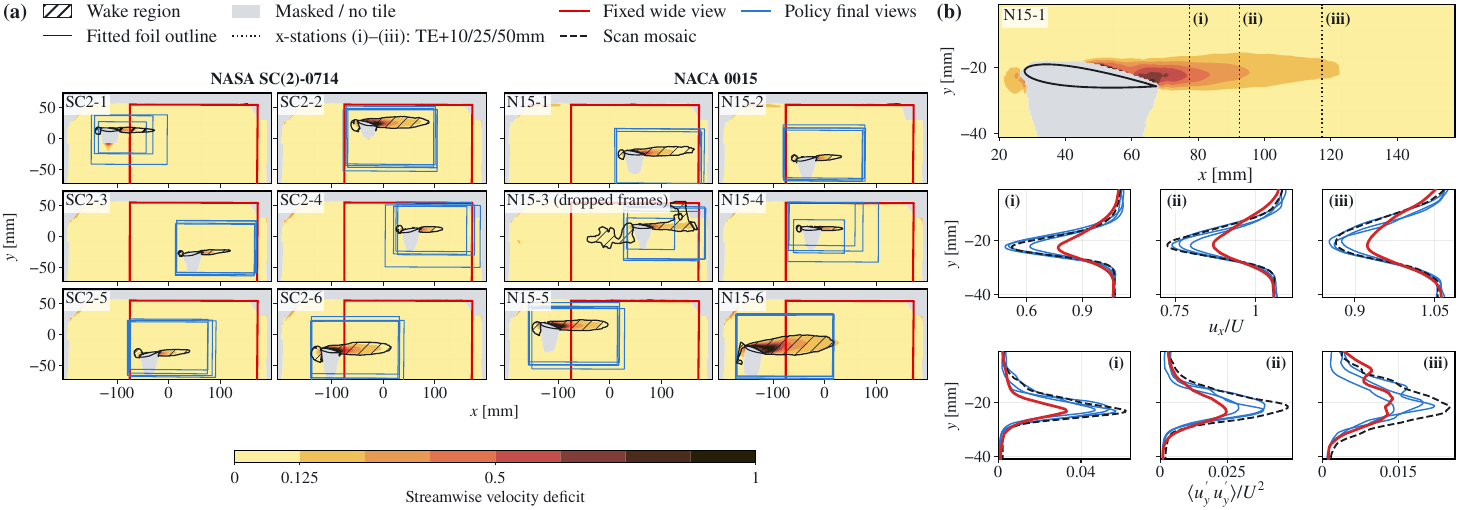}
    \caption{\textbf{Real policy deployments and flow statistics in the wake.} (a)~Velocity deficit from the scan mosaic for each hydrofoil setting, with the area imaged by the fixed wide view (red) and by the settled policy views (blue). The policy frames the whole wake in 41 of 42 views, whereas the fixed view cuts it in several settings. Setting N15-3 is shown but excluded from the comparison with the mosaic because the camera dropped frames during the scan. (b)~Setting N15-1. Profiles of $u_x/U$ and $\langle u_y'u_y'\rangle / U^2$ against $y$ at the stations (i)--(iii), which are 10, 25 and 50\,mm behind the trailing edge. The policy views (blue) follow the scan mosaic (dashed) more closely than the fixed wide view (red).}
  \label{fig:positions}
\end{figure*}

We transfer our policy zero-shot to hardware and assess whether it can locate the wake, adjust its framing when the wake changes, and improve flow-measurement accuracy in the real water tunnel. We evaluate it on two different 3D-printed hydrofoil profiles: a NACA 0015 and a NASA SC(2)-0714, neither included in training. The foils are mounted magnetically to the back wall of the water tunnel and extend to the front glass. Each hydrofoil is tested at six location-AoA combinations, labelled SC2-1 to SC2-6 and N15-1 to N15-6 in the order tested. For each setting, we run two 15-step episodes from different random camera starting positions. Then, a third episode continues after step 15: we move the hydrofoil to the next setting and let the policy run for another 30 steps without resetting the episode. The actual measurement is taken from the final viewpoint, from which we record 100 frame pairs and average their flow estimates; all results on hardware are computed from these fields. 

We compare the policy with a fixed-view baseline that observes most of the test section and we use a stitched mosaic as the reference reconstruction. The mosaic is obtained by scanning the full water tunnel in 15 close tiles, each comprising 100 image pairs, and stitching them in post-processing. The fixed view is taken from the farthest distance reachable by the arm and covers about 260 mm of the 420 mm test section. Covering the entire section would require a wider lens or a greater distance, both of which would further reduce wake resolution. All hydrofoil settings and camera start positions were selected randomly before the experiments. 

Fig.~\ref{fig:positions} shows that the policy converges to nearly the same view for each hydrofoil configuration from different initial positions, typically within 4\,mm. In all but one episode, it settles on a view containing the whole wake, even when the hydrofoil is moved mid-episode. A full 15-step episode, including the 100 frame pairs recorded at the final viewpoint, took a median of 85\,s, against 148\,s for the scan to cover the test section of our small water tunnel with 15 overlapping close-up views, and the policy processed 160 image pairs instead of the scan's 1500. This gap depends on the camera field of view as well as tunnel size, and would be far larger in full-scale water and wind tunnels. The policy captures the wake within a median of two steps, both from a random initial view and after a hydrofoil change, comparable to simulation. We encourage the reader to see the supplementary video, which shows the search, the settled views and the response to a hydrofoil change on the real setup. By comparison, the fixed view misses part of the wake in a third of the settings, despite imaging the sheet at roughly two thirds of the resolution achieved by the policy. We report every pre-planned run as recorded, without repeating any. Therefore, setting N15-3, in which the camera dropped frames during the mosaic scan, is shown in Fig.~\ref{fig:positions} but excluded from comparisons with the mosaic. Table~\ref{tab:rel_error_metrics} summarises, across the remaining eleven settings, the number of views containing the whole wake, the median distance between view centres reached from different initial positions, and the median pixel size on the light sheet.

To assess flow-measurement accuracy, we follow PIV studies of wake behaviour~\cite{Lee2008} and comparisons of PIV techniques~\cite{Willert2023}. For one case, Fig.~\ref{fig:positions} shows the non-dimensionalised, time-averaged streamwise velocity profiles $u_x/U$ and transverse velocity fluctuations $\langle u_y'u_y'\rangle/U^2$ at three streamwise stations in the wake. Both quantities are closer to the scan mosaic values at the policy's final view than at the baseline position. We quantify this improvement using $q \in \{\delta, u_x/U, \langle u_y'u_y'\rangle/U^2\}$ and calculate the relative summed error $\epsilon(q)$ and signed bias $\beta(q)$ of $q$ with respect to the scan:
\begin{equation}
  \begin{gathered}
    \epsilon(q) = \frac{\sum_{i \in \Omega} \left| q_i^{\mathrm{view}} - q_i^{\mathrm{scan}} \right|}
                       {\sum_{i \in \Omega} \left| q_i^{\mathrm{scan}} \right|}, \\[4pt]
    \beta(q) = \frac{\sum_{i \in \Omega} \left( q_i^{\mathrm{view}} - q_i^{\mathrm{scan}} \right)}
                    {\sum_{i \in \Omega} \left| q_i^{\mathrm{scan}} \right|}.
  \end{gathered}
\end{equation}
Here, $i$ denotes a cell within $\Omega$, the set of $0.5\,\mathrm{mm} \times 0.5\,\mathrm{mm}$ cells in the wake region that are valid (unmasked) in both the scan mosaic and the view. Table~\ref{tab:rel_error_metrics} reports the mean and standard deviation of both measures across the eleven hydrofoil settings, with the better value in bold. The policy clearly reduces both the error and the bias of the three flow-field variables within the wake region.

\begin{table}[h]
\centering
\footnotesize
\setlength{\tabcolsep}{4pt}
\caption{Fixed wide view against settled policy views on hardware.}
\begin{tabular}{c|ccc}
 & Wake in view & View spread [mm] & Pixel size [mm] \\
\hline
\rule{0pt}{2.5ex} Wide (fixed) & 7 of 11 & -- & 0.136 \\
Policy       & \textbf{41 of 42} & 3.6 & \textbf{0.090} \\[2pt]
\rule{0pt}{2ex} & $\epsilon(\delta)$ & $\epsilon(u_x/U)$ & $\epsilon(\langle u_y'u_y'\rangle/U^2)$ \\
\hline
\rule{0pt}{2.5ex} Wide (fixed) & 0.419 $\pm$ 0.212 & 0.075 $\pm$ 0.036 & 0.412 $\pm$ 0.150 \\
Policy       & \textbf{0.205 $\pm$ 0.049} & \textbf{0.044 $\pm$ 0.014} & \textbf{0.291 $\pm$ 0.099} \\[2pt]
\rule{0pt}{2ex} & $\beta(\delta)$ & $\beta(u_x/U)$ & $\beta(\langle u_y'u_y'\rangle/U^2)$ \\
\hline
\rule{0pt}{2.5ex} Wide (fixed) & -0.348 $\pm$ 0.248 & +0.064 $\pm$ 0.043 & -0.366 $\pm$ 0.196 \\
Policy       & \textbf{+0.006 $\pm$ 0.088} & \textbf{+0.001 $\pm$ 0.020} & \textbf{-0.066 $\pm$ 0.314} \\
\end{tabular}
\label{tab:rel_error_metrics}
\end{table}

\section{Limitations and Conclusion}

In this paper we present a method for autonomous PIV in which a camera mounted on a robotic arm improves measurement quality by locating and better resolving regions of interest in the measurement plane. The method is trained entirely in a photorealistic Isaac Sim environment that reproduces key visual artefacts of the real setup, including occlusions and shadows, and is deployed zero-shot on physical hardware. In hardware experiments, the policy located and framed the wakes of two hydrofoils not seen during training, both from random initial positions and after the hydrofoil was moved mid-episode. It settled on a view containing the whole wake in all but one episode. The robot, camera, lens, light source and computer used in the experiments cost approximately \texteuro15,000, roughly one tenth of the price of a commercial laser-based PIV system.

We expect the method to generalise to other wind- and water-tunnel setups, provided accurate CAD models of the scene are available, the lighting is adequately represented in simulation, and the camera is extensively calibrated. A limitation of the current work is that the agent does not adjust imaging parameters such as exposure time, inter-frame time and aperture, which also affect flow-reconstruction quality. Extending the policy to control these parameters is a natural direction for future work.

% ============================================================
%Acknowledgments to people or funding agencies should only be included after acceptance of the manuscript.
% ============================================================
\ifarxiv
\section*{Acknowledgements}
This work was supported by the Swiss National Science Foundation (SNSF)
under grant no.~10005849 (AuRoS). We thank Roger Vonbank for his valuable engineering contributions.
\fi

\bibliographystyle{IEEEtran}
\bibliography{references}

\end{document}